# Standard Fingerprint Databases
## Manual Minutiae Labeling and Matcher Performance Analyses

Mehmet Kayaoglu, Berkay Topcu, Umut Uludag

*TUBITAK – BILGEM, Informatics and Information Security Research Center, Turkey*

*{mehmet.kayaoglu, berkay.topcu, umut.uludag}@tubitak.gov.tr*
*www.ekds.gov.tr/bio*

## Abstract

*Fingerprint verification and identification algorithms based on minutiae features are used in many biometric systems today (e.g., governmental e-ID programs, border control, AFIS, personal authentication for portable devices). Researchers in industry/academia are now able to utilize many publicly available fingerprint databases (e.g., Fingerprint Verification Competition (FVC) & NIST databases) to compare/evaluate their feature extraction and/or matching algorithm performances against those of others. The results from these evaluations are typically utilized by decision makers responsible for implementing the cited biometric systems, in selecting/tuning specific sensors, feature extractors and matchers. In this study, for a subset of the cited public fingerprint databases, we report fingerprint minutiae matching results, which are based on (i) minutiae extracted automatically from fingerprint images, and (ii) minutiae extracted manually by human subjects. By doing so, we are able to (i) quantitatively judge the performance differences between these two cases, (ii) elaborate on performance upper bounds of minutiae matching, utilizing what can be termed as "ground truth" minutiae features, (iii) analyze minutiae matching performance, without coupling it with the minutiae extraction performance beforehand. Further, as we will freely distribute the minutiae templates, originating from this manual labeling study, in a standard minutiae template exchange format (ISO 19794-2), we believe that other researchers in the biometrics community will be able to utilize the associated results & templates to create their own evaluations pertaining to their fingerprint minutiae extractors/matchers.*

*Keywords– Authentication, dactyloscopy, database, extractor, fingerprint, Fingerprint Verification Competition (FVC), ISO 19794-2, labeling, matcher, minutia, NFIQ, performance, quality, ROC, sensor, standard, template.*

## 1. Introduction

Biometric systems that use physiological and/or behavioral characteristics (e.g., fingerprint, face, iris, speech, handwriting) for personal authentication are becoming ubiquitous: from border control to accessing smart phones, from employment applicants' criminal background checks to accessing controlled substances in hospitals, biometric technology is being deployed at a rapid rate. The inability of other authentication mechanisms that are based on either possession (e.g., ID cards, keys) and/or knowledge (e.g., passwords, PINs) in detecting imposters (e.g., individuals who gain control of other persons' cards and/or passwords), and in providing an irrefutable credential ("someone stole my card", "they possibly guessed my password"), along with the onset of practical, cheap(er), accurate, small, and durable sensors, can be among the reasons for the wide spread adoption of biometric technology. Even though there are many issues surrounding biometrics that still await careful analyses and, hopefully, solutions (e.g., sensor interoperability, user privacy, template protection, efficient database search), it seems "who you are" is and will be the preferred question, over/in addition to "what you have" and "what you know" in the foreseeable future.

Within many biometric modalities available, fingerprint-based systems account for the largest market share [1] and affect more people. The reasons for this preference could include (i) relatively more practical size & cost of fingerprint sensors compared to others, (ii) high authentication accuracy, and (iii) fingerprint template standards (e.g., ISO 19794-2 [2]) that allow sensor/algorithm interoperability. Furthermore, for fingerprint biometric, there are many standardized tests (e.g., [3-6], [7], [8]) and public databases (e.g., [3-6]) that help decision makers in government/industry/academia in objectively evaluating sensor &

algorithm performances. Utilizing existing databases also economically makes sense, as it frees the time and monetary resources that would otherwise be spent on collecting data, with the required (and generally quite demanding) characteristics (e.g., representative demographic distribution, size large enough to guarantee statistical significance of the associated performance results [9]). Existing fingerprint-based AFIS (Automated Fingerprint Identification System) infrastructure of law-enforcement agencies (e.g., FBI, police, homeland security) also contribute to this popularity.

Among the cited public fingerprint databases, FVC (Fingerprint Verification Competition [3-6]) images became a popular choice for researchers/engineers in both academia and industry (e.g., [10-13], [14-15]). The reason for this may be the relatively large size of these databases, utilization of many sensors & different technologies (e.g., optical, capacitive, thermal) for fingerprint image capture, and the participation of many academic and industrial algorithms (with a possibility of staying anonymous) in the associated standardized tests. For this reason, in this study, we concentrated on FVC fingerprint databases as our test beds. As will be explained later in greater detail, utilizing a GUI tool that is developed in our laboratory, we manually marked & extracted minutiae features (the most popular fingerprint feature) using four (FVC2002 DB1A-DB3A and FVC2004 DB1A-DB3A) of these databases' fingerprint images.

Utilizing two different commercial (automatic) minutiae matchers (that participated in cited FVC competitions), we obtained minutiae matching results pertaining to manually & automatically (via the commercial minutiae extractors associated with the cited matchers) extracted minutiae sets. A statistically significant performance improvement (which is quite logical to expect) with the manually extracted minutiae is observed. Note that manual minutiae extraction is also reported to improve performance for the latent fingerprint images as well [16]. To the best of our knowledge, our study is the first one to deal with manually labeled FVC databases. Note that Busch *et al.* [17] utilized dactyloscopic experts from a law-enforcement agency (German Federal Criminal Police) for marking minutiae features of a subset of NIST SD14 and SD29 databases (mostly ink-on-paper based acquisition), for minutiae conformance testing. In [18], authors utilized a (non-public) "ground truth" minutiae database, in order to evaluate minutiae statistics for biometric cryptosystem applications. Note that, NIST Special Database 27 [19] contains latent images from crime scenes and rolled on paper/scanned images with marked minutiae data. But this database does not provide the civil access control scenario characteristics (in terms of imaging/acquisition methodology & database size) that are needed for our study.

We are also able to quantify the improvement based on the underlying fingerprint acquisition methodology (as two of the utilized databases are obtained with optical sensors, one with a capacitive sensor, and one with a thermal sweeping sensor). Detailed analyses regarding the number of extracted minutiae, fingerprint image quality, and ROC (Receiver Operating Characteristic) curves allowed us to reach a performance upper bound, tied to the manually extracted "ground truth" minutiae features. Further, these analyses led to the possibility of evaluating solely minutiae *matching* performances, without incorporating the minutiae *extraction* performances.

The rest of the paper is organized as follows: in Section 2, we briefly summarize the characteristics of the GUI tool (*Fingerprint Manual Minutia Marker - FM3*) that we developed and utilized for minutiae marking. In Section 3, we provide information about the manually marked fingerprint databases and the marking procedure. Section 4 contains detailed analysis results pertaining to minutiae matching accuracy and other relevant metrics for manual and automatic minutiae extraction. Section 5 concludes the paper and provides pointers for future work.

## 2. Fingerprint Manual Minutia Marker – FM3

The tool that we developed and used for manually marking and extracting minutia features from fingerprint images is shown in Figure 1 (before and after marking an image). Using this interface, human subjects marked: (i) fingerprint image quality (poor/fair/good, as perceived by him/her), (ii) singular point (core and delta) locations, and (iii) minutiae data: for every minutia that he/she can identify, its type (ending/bifurcation), its location in image coordinates, its angle, and its quality (poor/fair/good). Note that we have utilized definitions of minutiae features (ie. location, angle conventions) as given in ISO/IEC 19794-2:2005 standard [2], to arrive at a consistent and widely used (in academia & industry) representation. Fingerprint type (arch/left loop/right loop/whorl) and completeness (poor/fair/good) are not utilized for the current study.

After marking is completed, the interface saves the extracted information into an ISO 19794-2 compliant template record (e.g., 1_1.iso-fmr), associated with the source image. Note that, the GUI window was sized so that all the human subjects saw every fingerprint image with a constant height of 22 centimeters on his/her computer monitor during manual marking.

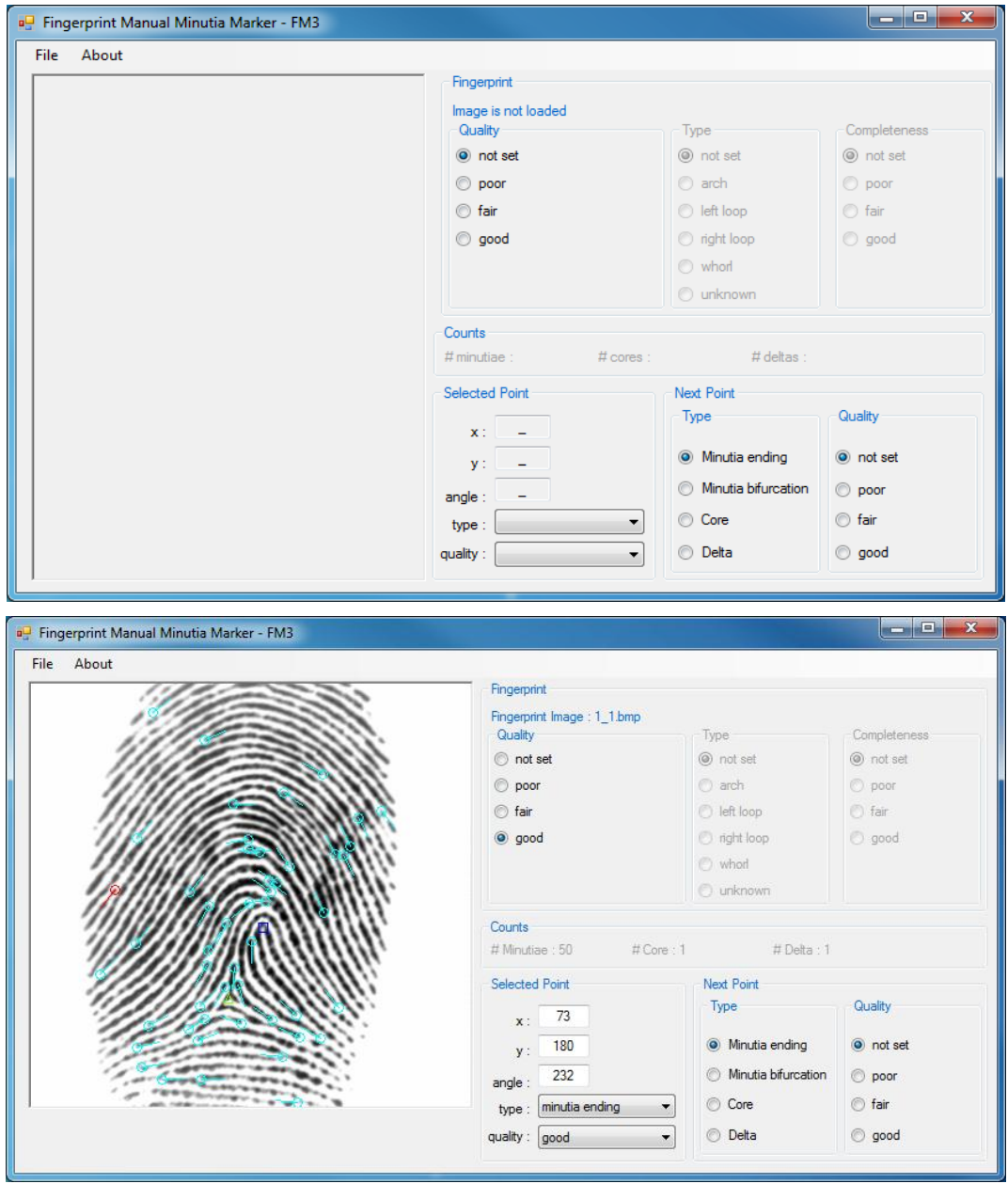

**Figure 1.** Fingerprint Manual Minutia Marker (FM3) graphical user interface.

## 3. Manually Marked Minutia Databases

With the FM3 tool summarized in Section 2, this manuscript's authors performed the manual feature marking of all the fingerprint images of the following four FVC databases:

- **FVC2002 DB1A**: Total number of images = 800; 100 unique fingers, 8 impressions per finger; 388x374 grayscale images captured with a 500 DPI **optical** sensor.

- **FVC2002 DB3A**: Total number of images = 800; 100 unique fingers, 8 impressions per finger; 300x300 grayscale images captured with a 500 DPI **capacitive** sensor.

- **FVC2004 DB1A**: Total number of images = 800; 100 unique fingers, 8 impressions per finger; 640x480 grayscale images captured with a 500 DPI **optical** sensor.

- **FVC2004 DB3A**: Total number of images = 800; 100 unique fingers, 8 impressions per finger; 300x480 grayscale images captured with a 512 DPI **thermal sweeping** sensor.

Sample images from these databases are shown in Figures 2-5 (shown image sizes are to scale; all images are from finger id 1, in respective databases). The 4 human subjects (including the paper authors, 3 males & 1 female, age range: 25 - 36) that performed the marking were not dactyloscopic experts; rather, they were electronics & computer engineers, with an average of 7 years of *automatic* fingerprint processing / algorithm development / tuning experience (combined 28 years, PhD and/or governmental e-ID project –including fingerprint biometric– work).

Each one of the cited four databases was split into 4 disjoint sets with 200 images each. With this split, every subject was given 2 images corresponding to every one of 100 unique fingers in a database. Every subject, without consulting the other subjects in any way, completed the manual marking of his/her 200 images in approximately 3 weeks. On average, 14 images per work day were marked by a subject; marking of a single image took, on average, 12 minutes (he/she spent, approximately 3 hours per work day in marking images). After this marking procedure is completed for a database, to eliminate any potential human errors introduced during that stage (e.g., missing minutia, spurious minutia, and errors in minutia coordinates / angles / quality), every marked image & associated templates are checked using the marking tool, and if necessary, templates are modified, by the 3 human subjects, that have not originally marked the aforementioned image. After this post processing, with the final minutiae templates, the analysis results shown in Section 4 were obtained (note that these final ISO 19794-2 template files will be distributed freely at our laboratory web site). Totally for the four databases, close to 116.000 minutiae were marked by human subjects.

Note that, the subjects, at any given time, could look at only a *single* fingerprint image and they were not allowed to look at multiple impressions of the same finger, during minutiae marking. This way, it was possible to generate a platform where a fair comparison with an *automatic* minutiae extractor (that also, generally, looks at a single image to extract the corresponding minutiae) could be made. Further, the marking schedule was arranged so that a subject could not see the image pair corresponding to the same unique finger in the same day, to prevent any biases.

## 4. Experimental Results

In this section, we analyze the authentication performance differences between manually extracted and automatically extracted minutiae scenarios, utilizing metrics such as minutiae count, matching score, and fingerprint image quality distributions. All of the 800 images (and associated minutiae templates) for a database are used, as necessary, for the following analyses. Note that, both of the commercial minutia extractor/matcher systems (denoted as "Extractor 1 - E1", "Matcher 1 - M1" and "Extractor 2 - E2", "Matcher 2 - M2") utilized just the minutiae coordinate/angle information (ie. they have not used the extended minutiae data, and they have not used the singular point data). Hence, we were able to quantify just the minutiae matching accuracy differences in our analyses given below. For example, Figure 6 shows a sample fingerprint image with overlaid minutiae.

Further, even though we are not disclosing the vendor ID's of these commercial systems, we would like to note that they were among the best (in terms of authentication accuracy) 20% of the participants among the associated FVC competitions. We are aware that, the SDK's that we have recently obtained from these vendors, possibly contain improvements in minutiae extraction / matcher algorithms, with respect to their original submissions, as these competitions were held years ago. We also understand that, the results that we present here are tied to the specific commercial systems that we utilized, but we also believe that they can provide pointers for other minutiae-based systems as well.

Figure 7 shows the minutiae count distributions and Table 1 summarizes the statistics for these distributions. We can see that, generally, Extractor 1 results in more minutiae than Extractor 2. Also, for the optical fingerprint databases, human subjects and automatic minutia extractors has resulted in similar minutiae distributions. On the other hand, especially for the thermal database, automatic extractors led to significantly higher number of minutiae. This may be attributed to the noisy nature of this database's images, leading to many spurious minutiae for the automatic extraction scenarios.

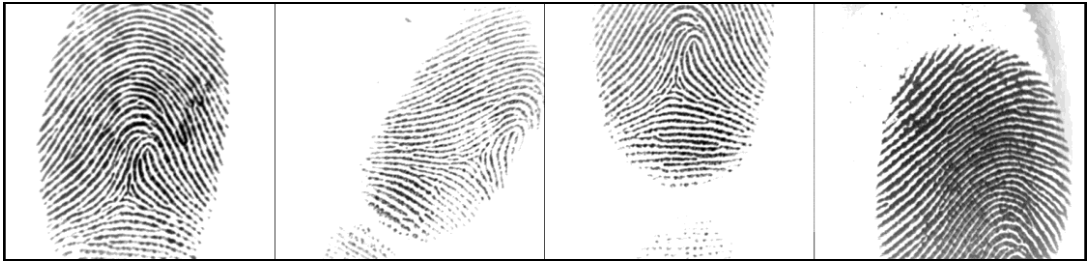

**Figure 2.** Images from FVC2002 DB1A database (optical).

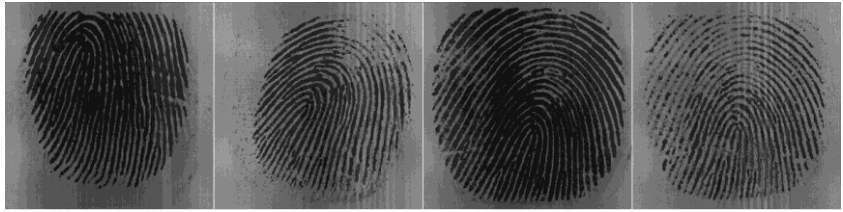

**Figure 3.** Images from FVC2002 DB3A database (capacitive).

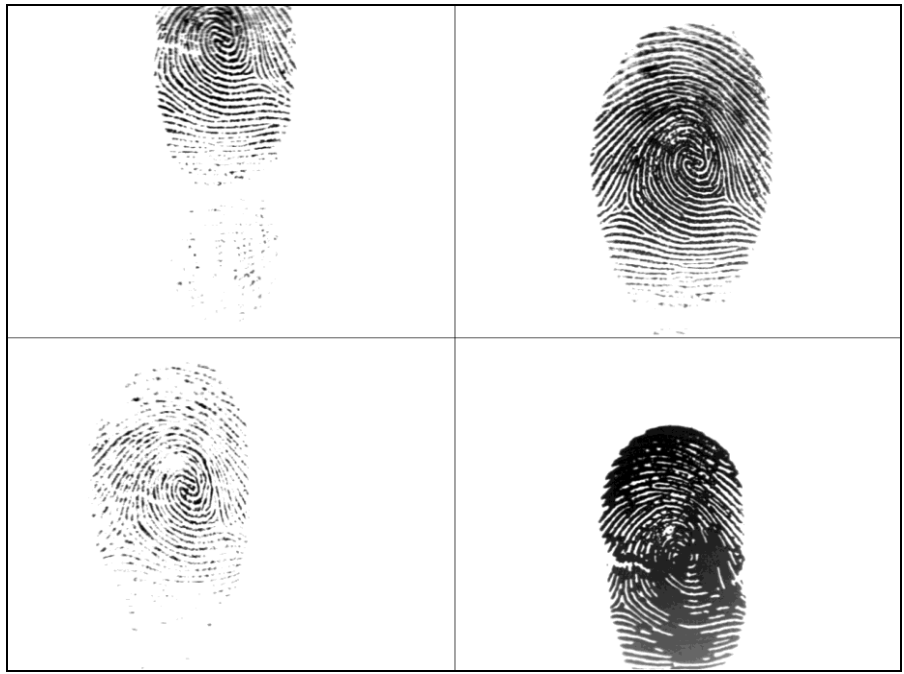

**Figure 4.** Images from FVC2004 DB1A database (optical).

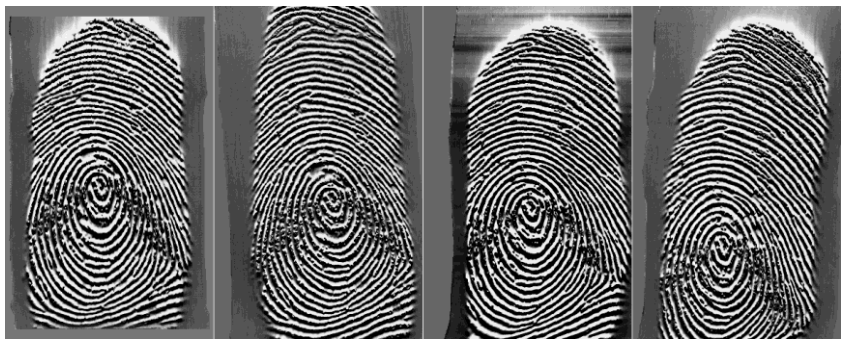

**Figure 5.** Images from FVC2004 DB3A database (thermal).

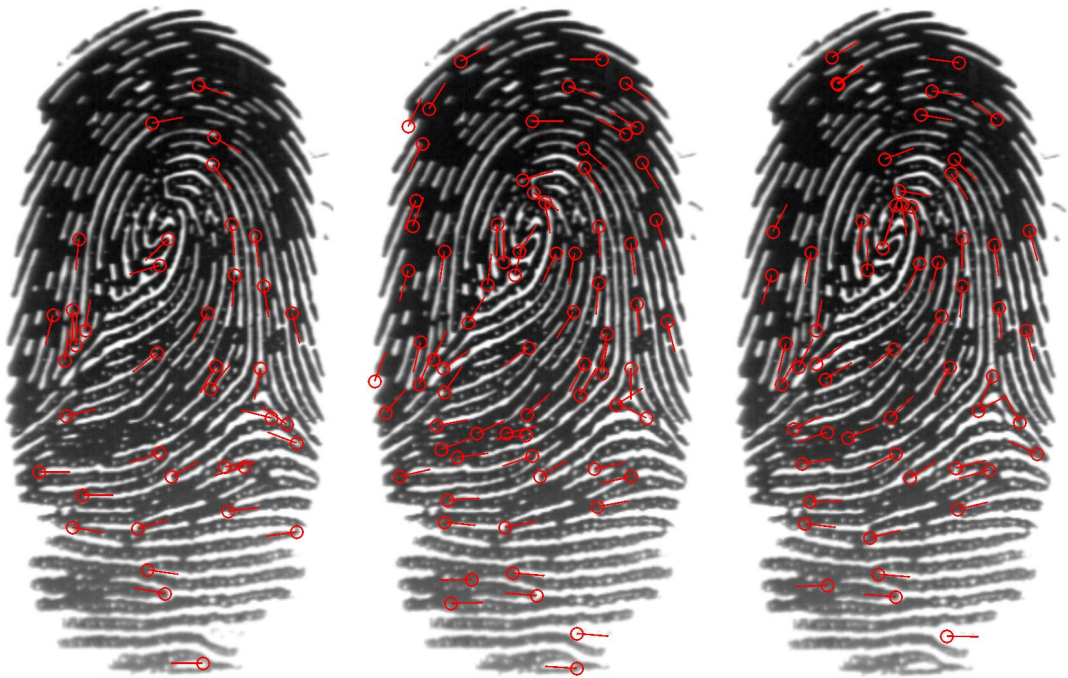

**Figure 6.** Sample image from FVC2004 DB1A database (optical, impression from finger id 12).
Left-to-right: manually extracted minutiae (count: 39), minutiae extracted with Extractor 1 (count: 72),
minutiae extracted with Extractor 2 (count: 58).

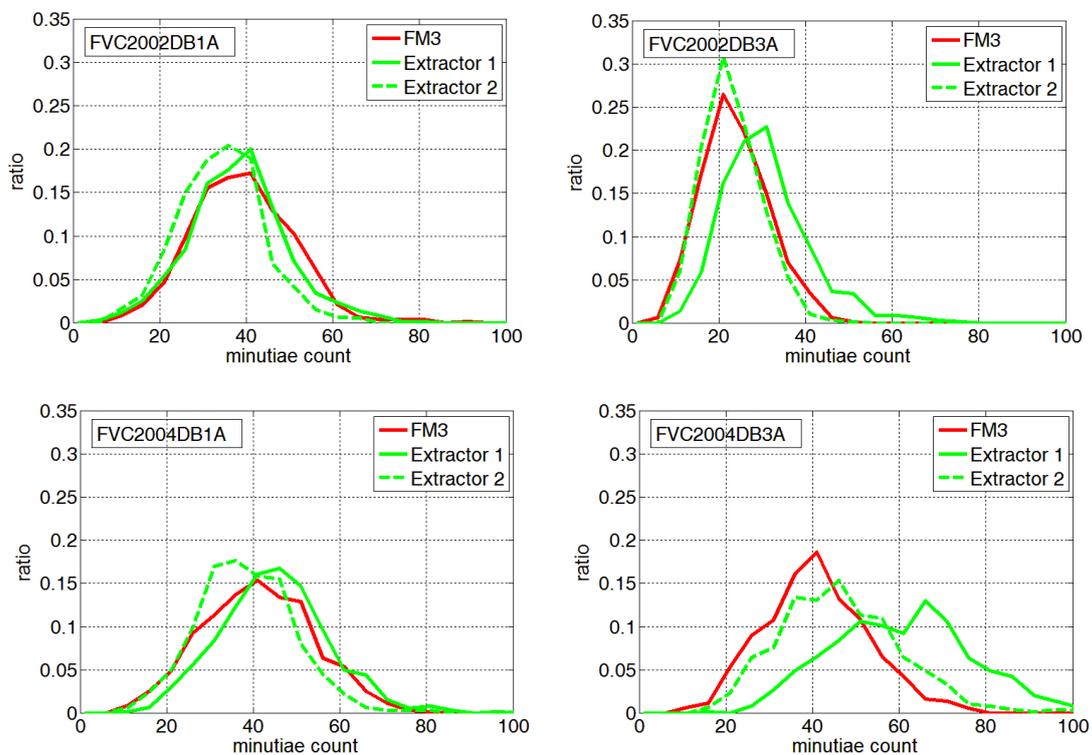

**Figure 7.** Minutiae count distributions for the databases:
manual marking (FM3) vs. automatic extraction.

**Table 1.** Minutia count statistics for the databases: manual marking (FM3) vs. automatic extraction.

| FVC2002DB1A | Mean | Std | Min | Max |
|---|---|---|---|---|
| FM3 | 39,1 | 11,4 | 9 | 92 |
| Extractor 1 | 38,1 | 11,2 | 5 | 80 |
| Extractor 2 | 34,0 | 9,7 | 8 | 68 |

| FVC2002DB3A | Mean | Std | Min | Max |
|---|---|---|---|---|
| FM3 | 23,8 | 7,6 | 6 | 49 |
| Extractor 1 | 30,9 | 10,0 | 9 | 74 |
| Extractor 2 | 22,6 | 6,8 | 6 | 50 |

| FVC2004DB1A | Mean | Std | Min | Max |
|---|---|---|---|---|
| FM3 | 41,0 | 12,6 | 11 | 80 |
| Extractor 1 | 45,1 | 12,8 | 12 | 116 |
| Extractor 2 | 38,4 | 11,3 | 10 | 101 |

| FVC2004DB3A | Mean | Std | Min | Max |
|---|---|---|---|---|
| FM3 | 40,8 | 11,9 | 11 | 76 |
| Extractor 1 | 62,8 | 19,7 | 16 | 163 |
| Extractor 2 | 48,0 | 16,9 | 14 | 132 |

Utilizing these commercial (automatic) minutiae matchers, that operated on manually & automatically extracted minutiae templates seperately, genuine and imposter matching scores are obtained. All of the possible genuine and imposter matches were performed, without any identical template match. This resulted in 5.600 genuine and 633.600 imposter matches, for a database, for all of the manual (FM3) and automatic extraction scenarios. In order to obtain a fair comparison between FM3 and automatic extraction, exactly the same matching pairs are used for these scenarios.

Figure 8 shows the resulting ROC curves. Considering these figures, along with Table 2, which lists GAR (Genuine Accept Rate) values and 95% confidence intervals for three FAR (False Accept Rate) values (ie. %0,001, %0,01 and %0,1), it can be observed that, manually extracted minutia scenario results in significantly higher authentication performance for all of the considered databases. The performance difference between FM3 and automatic scenarios widens even more for lower FAR values (higher security region of ROC). Further, for the capacitive and thermal databases, the performance differences are much stronger than those of optical databases. We should also note that, FVC 2004 database images contain the effects of exaggerated distortions introduced during fingerprint sensing, which lead to inferior authentication performances with respect to the FVC 2002 databases.

The results presented here can be used as a quantitative guide to see how much manually extracted minutiae ("ground truth") changes the fingerprint matching accuracy for the widely used FVC fingerprint databases.

For evaluating the effects of fingerprint image quality as perceived by human subjects (marked as poor/fair/good) and using automatic NFIQ algorithm of NIST [20], we report the associated distributions in Figures 9-10. Note that, the subjects evaluated the relative image quality, by considering the overall sensor-dependent image characteristics, along with the individual image characteristics (so, for example, the larger –with respect to the capacitive/thermal– and higher contrast images of optical database are not directly marked as "good quality", but their relative noise levels, presence of smudges, etc. are evaluated as well).

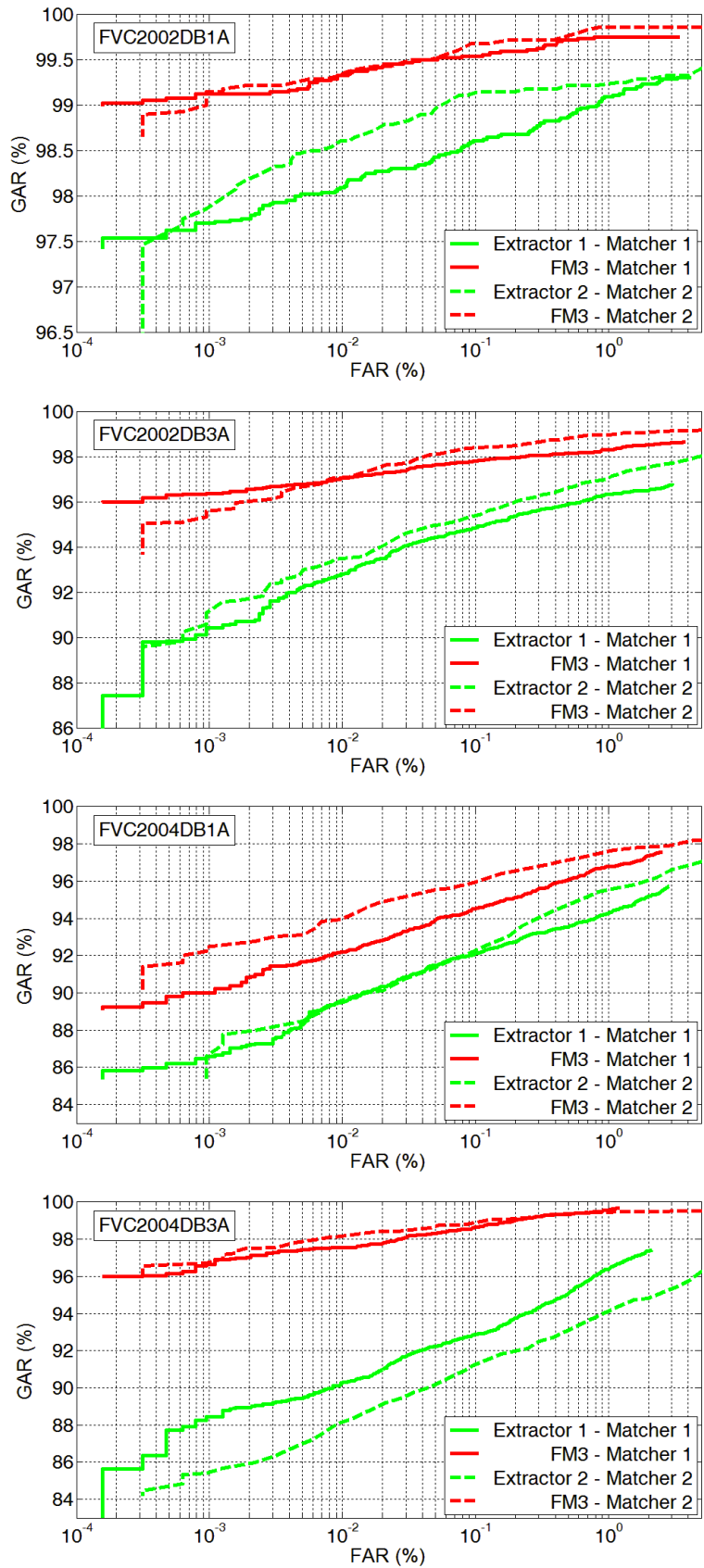

**Figure 8.** ROC's for analysed databases.

**Table 2.** GAR (%) for three FARs: Red: Value, Black: Confidence Interval.

| FVC2002DB1A | FAR (%) | | |
|---|---|---|---|
| | 0,001 | 0,01 | 0,1 |
| FM3 - M1 | [98,8 - 99,1 - 99,4] | [99,1 - 99,3 - 99,5] | [99,3 - 99,5 - 99,7] |
| E1 - M1 | [97,3 - 97,7 - 98,1] | [97,7 - 98,1 - 98,4] | [98,3 - 98,6 - 98,9] |
| FM3 - M2 | [98,9 - 99,1 - 99,4] | [99,1 - 99,4 - 99,6] | [99,5 - 99,7 - 99,8] |
| E2 - M2 | [97,4 - 97,8 - 98,2] | [98,3 - 98,6 - 98,9] | [98,9 - 99,1 - 99,4] |

| FVC2002DB3A | FAR (%) | | |
|---|---|---|---|
| | 0,001 | 0,01 | 0,1 |
| FM3 - M1 | [95,9 - 96,4 - 96,9] | [96,6 - 97,0 - 97,5] | [97,4 - 97,8 - 98,2] |
| E1 - M1 | [89,6 - 90,4 - 91,2] | [92,1 - 92,8 - 93,5] | [94,2 - 94,8 - 95,4] |
| FM3 - M2 | [95,0 - 95,6 - 96,1] | [96,6 - 97,1 - 97,5] | [98,0 - 98,4 - 98,7] |
| E2 - M2 | [90,3 - 91,1 - 91,8] | [92,8 - 93,5 - 94,1] | [94,9 - 95,5 - 96,0] |

| FVC2004DB1A | FAR (%) | | |
|---|---|---|---|
| | 0,001 | 0,01 | 0,1 |
| FM3 - M1 | [89,2 - 90,0 - 90,8] | [91,5 - 92,2 - 92,9] | [93,9 - 94,5 - 95,1] |
| E1 - M1 | [85,7 - 86,7 - 87,5] | [88,7 - 89,5 - 90,3] | [91,4 - 92,1 - 92,8] |
| FM3 - M2 | [91,7 - 92,5 - 93,1] | [93,3 - 94,0 - 94,6] | [95,4 - 95,9 - 96,4] |
| E2 - M2 | [85,7 - 86,6 - 87,5] | [88,7 - 89,5 - 90,3] | [91,6 - 92,3 - 93,0] |

| FVC2004DB3A | FAR (%) | | |
|---|---|---|---|
| | 0,001 | 0,01 | 0,1 |
| FM3 - M1 | [96,1 - 96,6 - 97,1] | [97,1 - 97,6 - 97,9] | [98,3 - 98,7 - 98,9] |
| E1 - M1 | [87,6 - 88,5 - 89,3] | [89,5 - 90,3 - 91,1] | [92,7 - 92,9 - 93,6] |
| FM3 - M2 | [96,2 - 96,7 - 97,2] | [97,7 - 98,1 - 98,5] | [98,5 - 98,8 - 99,1] |
| E2 - M2 | [84,4 - 85,4 - 86,3] | [87,2 - 88,1 - 88,9] | [90,4 - 91,2 - 91,9] |

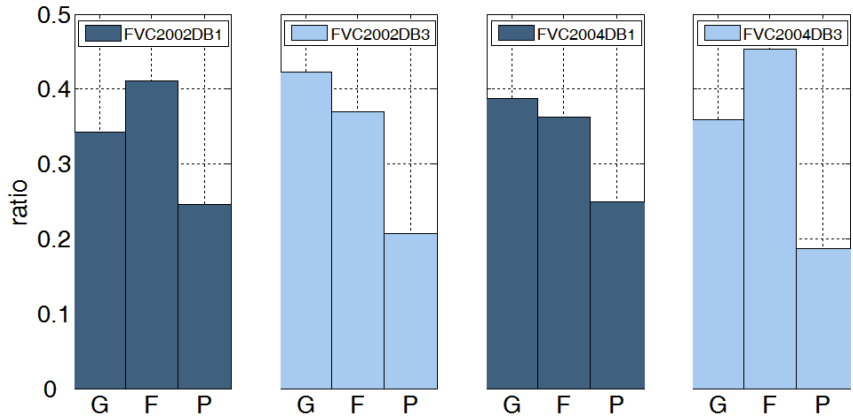

**Figure 9.** Fingerprint image quality (perceived by human subjects) distributions
for the databases (G: good, F: fair, P: poor).

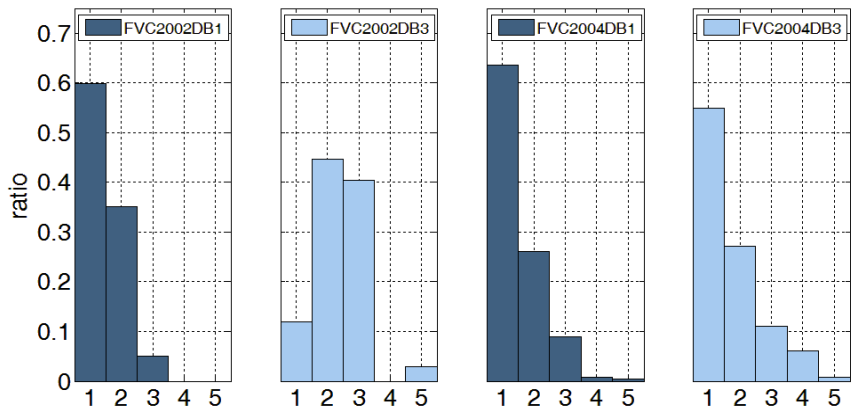

**Figure 10.** Fingerprint image quality (NFIQ scores) distributions for the databases (1: best, 5: worst).

In order to analyze the effects of a possible "rejection" option before minutiae matching, where "poor" quality images do not enter the matching phase at all, but only the "good" and "fair" quality images are matched, we report the associated ROC curves (Figure 11) and GAR vs. FAR values with 95% confidence intervals (Table 3). It can be observed that, for the thermal database, a larger improvement (e.g., more than % 2 GAR increase for % 0,001 FAR) is present. Note that, here our aim is to provide quantitative results and confidence intervals for this –expected– authentication performance improvement.

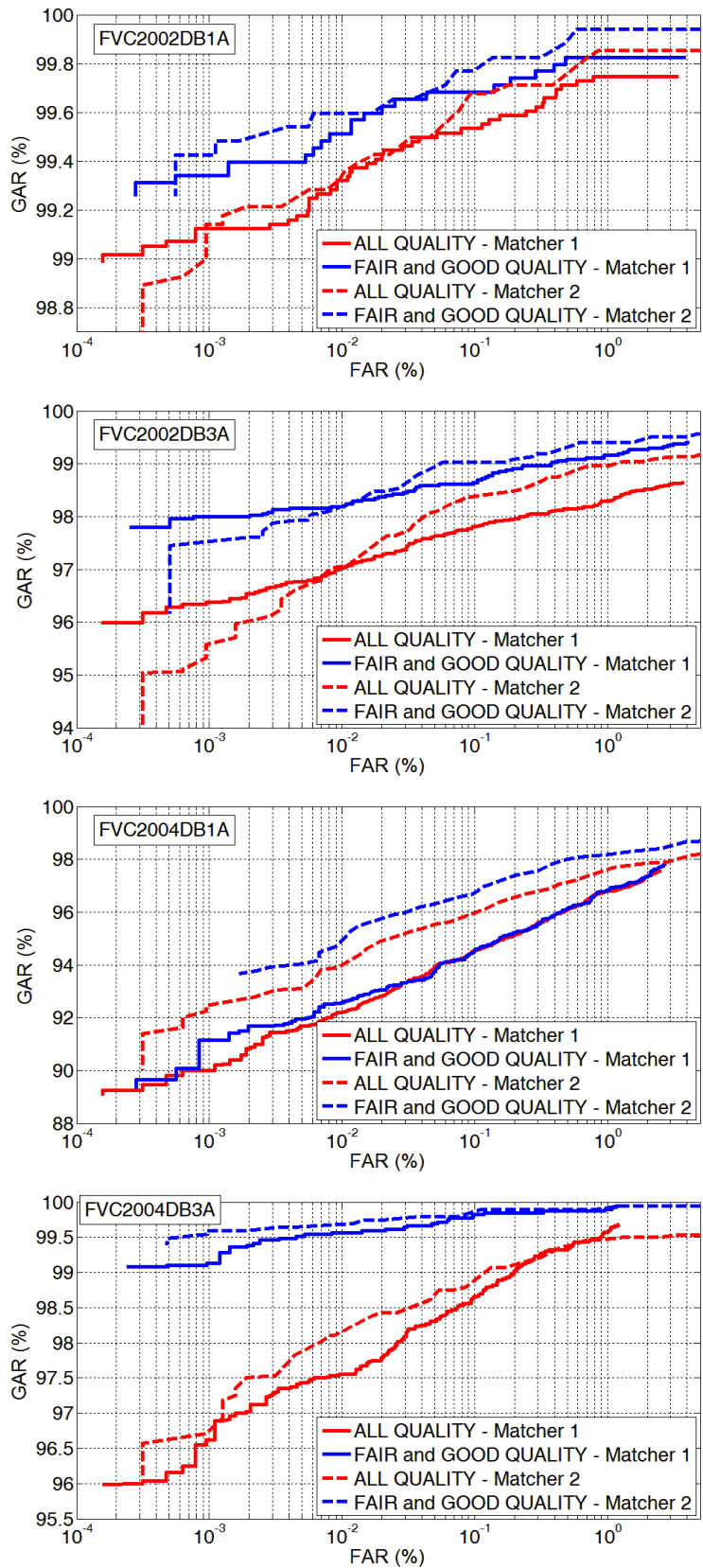

**Figure 11.** ROCs: All images vs. only images that are not marked as "poor"
(top-to-bottom: approx. %24, %20, %25, %18 of images are rejected, respectively).

**Table 3.** GAR (%) for three FARs (with and without image quality-based rejection): Red: Value, Black: Confidence Interval.

| FVC2002DB1A | FAR (%) | | |
|---|---|---|---|
| **FM3** | 0,001 | 0,01 | 0,1 |
| All Quality – M1 | [98,8 - 99,1 - 99,4] | [99,1 - 99,3 - 99,5] | [99,3 - 99,5 - 99,7] |
| Fair + Good – M1 | [99,0 - 99,3 - 99,6] | [99,2 - 99,5 - 99,7] | [99,4 - 99,7 - 99,8] |
| All Quality – M2 | [98,9 - 99,1 - 99,4] | [99,1 - 99,4 - 99,6] | [99,5 - 99,7 - 99,8] |
| Fair + Good – M2 | [99,2 - 99,5 - 99,7] | [99,3 - 99,6 - 99,8] | [99,6 - 99,8 - 99,9] |

| FVC2002DB3A | FAR (%) | | |
|---|---|---|---|
| **FM3** | 0,001 | 0,01 | 0,1 |
| All Quality – M1 | [95,9 - 96,4 - 96,9] | [96,6 - 97,0 - 97,5] | [97,4 - 97,8 - 98,2] |
| Fair + Good – M1 | [97,5 - 98,0 - 98,4] | [97,7 - 98,2 - 98,6] | [98,2 - 98,6 - 99,0] |
| All Quality – M2 | [95,0 - 95,6 - 96,1] | [96,6 - 97,1 - 97,5] | [98,0 - 98,4 - 98,7] |
| Fair + Good – M2 | [96,9 - 97,5 - 97,9] | [97,6 - 98,1 - 98,5] | [98,7 - 99,0 - 99,3] |

| FVC2004DB1A | FAR (%) | | |
|---|---|---|---|
| **FM3** | 0,001 | 0,01 | 0,1 |
| All Quality – M1 | [89,2 - 90,0 - 90,8] | [91,5 - 92,2 - 92,9] | [93,9 - 94,5 - 95,1] |
| Fair + Good – M1 | [90,1 - 91,2 - 92,1] | [91,6 - 92,6 - 93,5] | [93,7 - 94,5 - 95,3] |
| All Quality – M2 | [91,7 - 92,5 - 93,1] | [93,3 - 94,0 - 94,6] | [95,4 - 95,9 - 96,4] |
| Fair + Good – M2 | [92,8 - 93,7 - 94,5] | [93,8 - 94,7 - 95,4] | [96,0 - 96,7 - 97,3] |

| FVC2004DB3A | FAR (%) | | |
|---|---|---|---|
| **FM3** | 0,001 | 0,01 | 0,1 |
| All Quality – M1 | [96,1 - 96,6 - 97,1] | [97,1 - 97,6 - 97,9] | [98,3 - 98,7 - 98,9] |
| Fair + Good – M1 | [98,8 - 99,1 - 99,4] | [99,3 - 99,6 - 99,8] | [99,6 - 99,8 - 99,9] |
| All Quality – M2 | [96,2 - 96,7 - 97,2] | [97,7 - 98,1 - 98,5] | [98,5 - 98,8 - 99,1] |
| Fair + Good – M2 | [99,3 - 99,6 - 99,8] | [99,5 - 99,7 - 99,8] | [99,7 - 99,9 - 100] |

## 5. Conclusions and Future Work

Utilizing four popular FVC (Fingerprint Verification Competition) fingerprint image databases as our test beds, we manually marked minutiae features via a developed GUI tool. Comparing the minutiae matching accuracy for this scenario, with that of automatically extracted (via two commercial systems) minutiae, we were able to quantify the associated performance differences and observe a statistically significant improvement with the former. Optical, capacitive and thermal databases utilized in the study led to different magnitudes for the cited improvement, along with considerable automatic "spurious" minutiae extraction evidence, especially with the thermal database.

The performance figures (summarized as ROC curves and confidence intervals in the manuscript) may be regarded as "ground truth"- minutiae-based performance upper bounds, for the involved database / matcher platforms.

Also, it is found that, if it is possible to discard "poor" quality images (and utilize only the "fair" and "good" quality images that account for approximately % 80 of the total database size) during fingerprint matching, a further improvement in matching accuracy is achievable. We were able to quantify the magnitude of this expected result as well.

We will freely distribute the manually marked minutiae templates (that are are formatted according to a previously published ISO standard) for all of the associated databases in order to allow researchers in biometrics community to evaluate their fingerprint minutiae extractors / matchers.

Our future work will include analyzing the statistics of minutiae common to ("corresponding minutiae") automatic and manual extraction scenarios, and arriving at the comparison results for other databases' images. Further, these "ground truth" minutiae features can be utilized to arrive at –higher fidelity– fingerprint image deformation models, based on minutiae features.

## 6. Acknowledgements

We would like to thank Mrs. Merve Kilinc Yildirim for her help in manual minutiae marking studies summarized in the paper.